\documentclass[conference]{IEEEtran}
\IEEEoverridecommandlockouts
\usepackage{cite}
\usepackage{amsmath,amssymb,amsfonts}
\usepackage{algorithmic}
\usepackage{graphicx}
\usepackage{textcomp}
\usepackage{xcolor}

\usepackage{algorithm}
\usepackage{algorithmic}

\usepackage{array}
\usepackage{booktabs}
\usepackage{amsmath}
\usepackage{amssymb}
\usepackage{tcolorbox}

\usepackage{marvosym}
\usepackage{hyperref}

\usepackage{multirow}
\usepackage[normalem]{ulem}
\useunder{\uline}{\ul}{}

\def\BibTeX{{\rm B\kern-.05em{\sc i\kern-.025em b}\kern-.08em
    T\kern-.1667em\lower.7ex\hbox{E}\kern-.125emX}}
\begin{document}

\title{Refining Interactions: Enhancing Anisotropy in Graph Neural Networks with Language Semantics}

\author{
	Zhaoxing Li$^{1}$, Haifeng Zhang$^{2}$, Xiaoming Zhang$^{3,}$\textsuperscript{\Letter}, Chengxiang Liu$^{4}$ \\
        
	$^{1,3,4}$Institute of Physical Science and Information Technology, Anhui University \\
        $^{2}$School of Mathematical Sciences, Anhui University \\
        $^{3}$Qinghai Institute of Science and Technology Information, Xining, China\\
        
        \{$^{1}$lzx,$^{4}$Q23301281\}@stu.ahu.edu.cn,$^{2}$haifengzhang1978@gmail.com,$^{3,}$\textsuperscript{\Letter}xmzhang@ustc.edu
    
}
\maketitle
\renewcommand{\thefootnote}{}
\footnotetext{\textsuperscript{\Letter} Corresponding Author. 

\thanks{This work was supported by the Qinghai Key Research and Development Project(2024-SF-128).}
}

\begin{abstract}
The integration of Large Language Models (LLMs) with Graph Neural Networks (GNNs) has recently been explored to enhance the capabilities of Text Attribute Graphs (TAGs). Most existing methods feed textual descriptions of the graph structure or neighbouring nodes’ text directly into LLMs. However, these approaches often cause LLMs to treat structural information simply as general contextual text, thus limiting their effectiveness in graph-related tasks. In this paper, we introduce LanSAGNN (Language Semantic Anisotropic Graph Neural Network), a framework that extends the concept of anisotropic GNNs to the natural language level. This model leverages LLMs to extract tailor-made semantic information for node pairs, effectively capturing the unique interactions within node relationships. In addition, we propose an efficient dual-layer LLMs finetuning architecture to better align LLMs’ outputs with graph tasks. Experimental results demonstrate that LanSAGNN significantly enhances existing LLM-based methods without increasing complexity while also exhibiting strong robustness against interference.
\end{abstract}


\section{Introduction}
\label{sec:intro}

In recent years, graph learning has attracted widespread attention and research across various fields due to its outstanding performance in bioinformatics, chemistry, and social network analysis \cite{chen2024exploring}. Graph data extensively cover real-world scenarios, particularly those involving Text Attribute Graphs (TAGs). These graphs provide topological information between nodes and offer rich semantic information through text attributes.

TAGs contain textual attributes of nodes and structural information about the connections between them. In solving graph tasks, the textual attributes of TAGs are encoded by embedding algorithms into embedding vectors, which are then fed into graph neural networks (GNNs) \cite{zhao2023glem}. GNNs can utilize the structural information of TAGs through paradigms such as message passing to perform various downstream tasks. In traditional processes, shallow models represented by Bag of Words, Word2Vec, and TF-IDF are commonly used to encode the textual attributes of TAGs \cite{sparck1972statistical}. However, research has shown that these embedding models, due to their inability to understand the contextual information of texts effectively, limit the performance of downstream tasks to some extent \cite{miaschi2020contextual}.

With the recent emergence of Large Language Models (LLMs) in natural language processing, the graph learning community has also discovered that LLMs possess more powerful semantic understanding and contextual awareness than shallow embedding models \cite{chen2024exploring}. The use of LLMs as tools for representation learning in TAGs has been explored to some extent, and the success of these methods has demonstrated their tremendous potential in TAGs tasks. An intuitive approach uses carefully designed prompts to guide LLMs to produce higher-quality textual information and embeddings for TAGs, replacing traditional shallow embedding algorithms. As for the structural information of TAGs, it is processed by subsequent GNNs in the same manner as traditional workflows \cite{he2024harnessing}. Some researchers have also made progress by inputting combined node textual attributes and structural information into LLMs, demonstrating that structural information is equally crucial in LLM-based graph task paradigms \cite{guo2023gpt4graph}.

Recent explorations have found that feeding the graph structure or the original textual information of neighbours directly into LLMs has led to specific performance improvements. However, a recent study has shown that LLMs do not interpret this content as graph structures as intended by the designers but rather tend to treat it as regular contextual paragraphs \cite{Huang2023CanLE}. This implies that the existing integration models have not fully exploited the role of LLMs in handling the structural information within TAGs.

\begin{figure}[t]
\centering
\includegraphics[width=0.95\columnwidth]{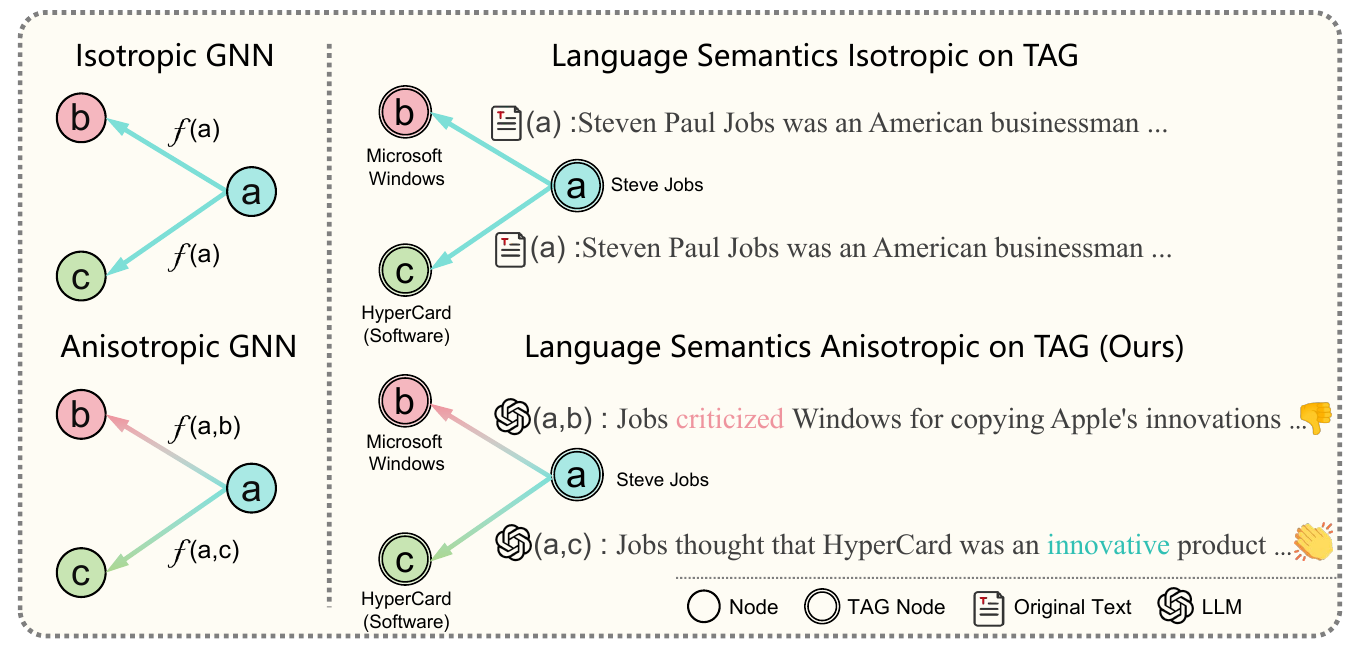} 
\caption{Traditional GNNs (left) versus Natural Language Level (right) isotropy and anisotropy.}
\label{Fig1}
\end{figure}

To address the aforementioned issues, we draw inspiration from anisotropic GNNs. In graph learning, the emergence of anisotropic networks is widely recognized as a significant progress \cite{dwivedi2023benchmarking}. Traditionally, anisotropy primarily manifests in the interactions of embeddings between nodes. We extend this concept from the embedding level to the natural language level. Fig. \ref{Fig1} vividly demonstrates the anisotropic features at both the embedding and natural language levels. The left side of the figure illustrates the message-passing in traditional GNNs. In isotropic models, the source node transmits the \textit{same} embedding to different nodes through the message-passing mechanism, whereas anisotropic GNNs deliver \textit{distinct} messages by interacting with different target nodes. The right side shows the differences between anisotropy and isotropy at the natural language level, with three nodes selected from the Wiki-CS dataset, the textual attributes of the nodes are the introductions of the nodes themselves. Isotropy in natural language semantics is evident as the original text of a node is directly utilized by neighbouring nodes. For example, in this case, both the \textit{Microsoft Windows} and \textit{HyperCard} (development software) nodes receive the same original text from the \textit{Steve Jobs} node. However, we extend anisotropy to the natural language semantics level through LLMs, enabling the LLMs to extract different attitudes of Steve Jobs towards Windows and HyperCard. In essence, the same central node, \textit{Steve Jobs}, ``customizes" different messages according to different target nodes. Compared to a personal introduction of \textit{Jobs}, his attitudes towards both are clearly more meaningful for the \textit{Microsoft Windows} and \textit{HyperCard} nodes.

Motivated by the aforementioned considerations, we propose LanSAGNN (\textbf{Lan}guage \textbf{S}emantics \textbf{A}nisotropic \textbf{G}raph \textbf{N}eural \textbf{N}etwork). Unlike existing integration models, this method does not directly encode the structural information or the original text of neighbors around TAGs' nodes to solve downstream tasks. Instead, it leverages the semantic understanding capabilities of LLMs to extract distinct contextual information for different node pairs within TAGs. This ensures that a node delivers closely related but distinct information to its various neighbors, rather than indiscriminately using raw text. This approach means that LLMs do not directly access all structural information during inference, thereby avoiding the problem of treating structural information as conventional context. Instead, complete structural information is utilized for aggregation within the method’s architecture.

A significant challenge is ensuring that the output from LLMs provides practical value for graph-related tasks. For small-scale language models, research \cite{ribeiro-etal-2021-structural} has proposed inserting multiple trainable GNNs adapter layers between the model layers. However, considering the inference speed and structural characteristics of LLMs, this approach is not feasible in practical applications. Therefore, we design an efficient dual-layer finetuning architecture tailored to the characteristics of graph data, which can quickly improve model performance with a small amount of data. Additionally, we design an edge filter that further enhances computational efficiency without sacrificing performance.

In summary, our contributions are as follows:

\begin{itemize} 
\item We extend the anisotropy in GNNs to the natural language level, introducing a new paradigm that combines LLMs with GNNs. 
\item We introduce an efficient dual-layer LLMs finetuning architecture tailored for graph data, allowing LLMs to adapt to graph tasks with low data and computational requirements and significantly enhancing performance.
\item Our proposed LanSAGNN achieves SOTA performance, consistently outperforming traditional GNNs and various LLM-based methods across multiple public datasets and LLMs without increasing complexity. \end{itemize}

\section{Related Work}
\subsection{Integration of LLMs and GNNs}
With the advent of LLMs, exploring how to enhance TAGs learning with LLMs has become a meaningful direction. One straightforward approach is to replace traditional shallow embedding models with LLM-based embedding models, addressing the issue of insufficient context capture in traditional embeddings. Additionally, some methods use LLMs as feature enhancers. For instance, TAPE \cite{he2023harnessing} guides LLMs through carefully designed instructions to output pseudo-label rankings and corresponding reasons for each node, achieving better results than directly embedding the original text.

Beyond merely applying LLMs to the text of individual nodes, some methods have started exploring how to better utilize the structural information of graph data through LLMs. Methods such as InstructGLM \cite{ye2023language} convert a node's neighbourhood structure into natural language text and pass it to LLMs. This process adapts LLMs to downstream tasks through finetuning and involves directly instructing the LLMs to produce results tailored to these tasks. GraphEdit \cite{guo2024graphedit} applies LLMs to graph structure learning by providing the original text of two nodes and finetuning the LLMs to determine whether an edge should exist between them, thus achieving excellent performance through optimized graph structures and LLM-enhanced embeddings.

\subsection{Isotropic and Anisotropic GNNs}
In the graph learning community, it is a common consensus that, in most cases, isotropic GNNs offer higher efficiency and lower memory overhead. In contrast, anisotropic GNNs trade off some complexity for higher accuracy \cite{tailor2022egc}. For example, GCN \cite{kipf2017semi} is a typical isotropic GNN, where the information passed in the message propagation process only relates to the source node itself. In contrast, GAT \cite{Liu_Zhou_2020} introduces an attention mechanism to facilitate interactions between source nodes and target nodes, making it a classic example of anisotropic GNN. We list more information about typical GNNs in the appendix, including  isotropic GNNs such as GIN \cite{xu2018how} and GraphSAGE \cite{hamilton2017inductive}, as well as anisotropic GNNs like MPNN \cite{10.5555/3305381.3305512} and PNA \cite{10.5555/3495724.3496836}.

The methods above primarily focus on utilizing node embeddings. However, message passing at the natural language level is equally crucial for TAGs. Therefore, we introduce LanSAGNN, a new model that extends the concept of anisotropy in graphs to the natural language processing level.

\begin{figure*}[t]
\centering
\includegraphics[width=1.5\columnwidth]{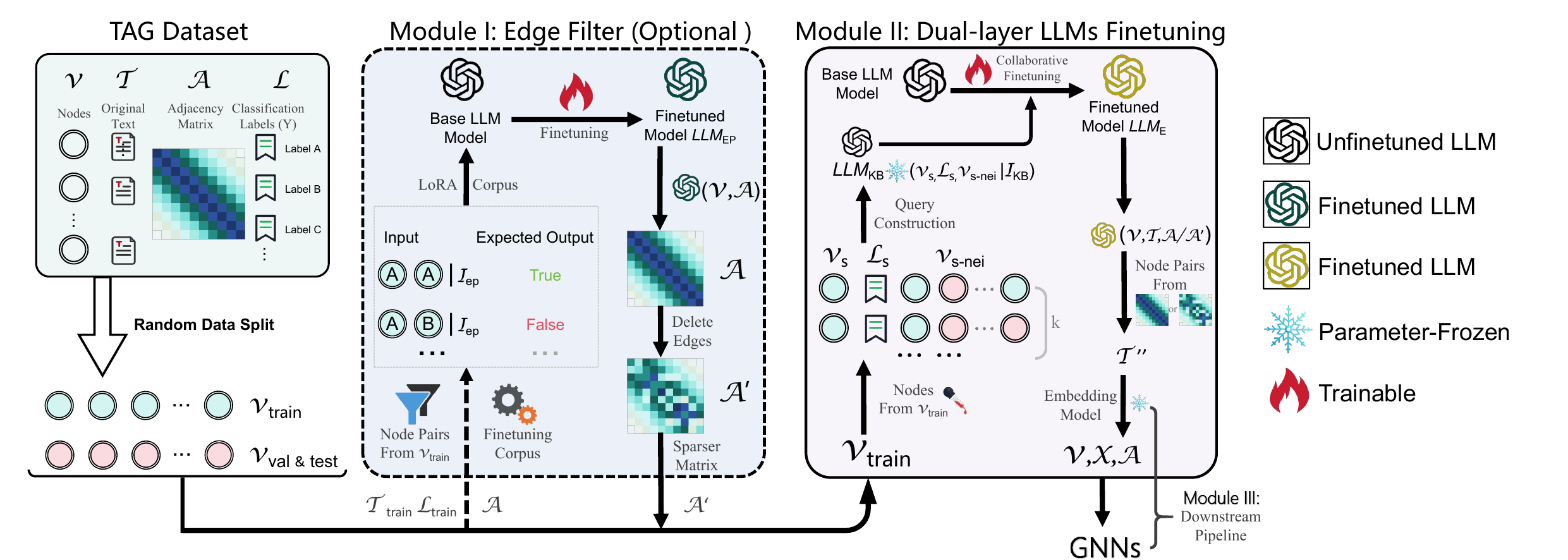} 
\caption{The overall framework of LanSAGNN.}
\label{Fig2}
\end{figure*}

\section{Method}
\subsection{Preliminaries}

Our method focuses on the node classification problem in TAGs dataset. We define a dataset as \( \mathcal{G} = (\mathcal{V}, \mathcal{A}, \mathcal{T}) \). Here, \( \mathcal{V} = \{v_1, \dots, v_n\} \) represents \( n \) nodes, each associated with an element \( t_i \) from the set of textual attributes \( \mathcal{T} \). \( \mathcal{A}  \in \mathbb{R}^{N \times N} \) is the adjacency matrix. Our task is to determine the class labels of classification task $\mathcal{Y}$ for all nodes, given the set of labels $ \mathcal{Y}_{\text{train}} $ for the training set.

\subsection{Overview of LanSAGNN}

As shown in Fig. \ref{Fig2}, LanSAGNN consists of the following modules: The optional edge filter module identifies whether node pairs are homogeneous based on their text attributes, and deleting heterogeneous node pairs enhances efficiency while ensuring performance. The dual-layer LLMs finetuning architecture module, through the cooperative action of two LLMs, compensates for the inability of LLMs to participate in end-to-end training with efficient finetuning, ensuring that the output aligns with downstream tasks and utilizes the structural information of the graph at the natural language level. Finally, the downstream pipeline module converts the output of the LLM into embeddings, which GNNs then process to complete downstream tasks, utilizing the structural information of the graph at the embedding level.

\subsection{Optional Edge Filter}
Anisotropic GNNs, compared to isotropic GNNs, have higher time complexity due to the need to consider interactions between each node and its neighbouring nodes. For each node, LanSAGNN requires extracting information based on its neighbourhood, which means that its time complexity is proportional to the number of edges. In large datasets, this results in low time efficiency. Therefore, LanSAGNN employs a filter to perform preliminary edge selection by deleting some edges to achieve higher efficiency. Extensive research indicates that homogeneity is an important source of performance for GNNs. By identifying and removing heterogeneous edges, it is possible to reduce computational demand while maintaining performance \cite{platonovcritical}. 

Specifically, several node pairs are randomly sampled among training set nodes. For any node pair: $v_{i},v_{j} \in \mathcal{V}_{train}$, and the original text $t_{i}$ and $t_{j}$ of the two nodes are combined through the instruction $I_{\text{EP}}$ as input for finetuning, with the output based on whether the classification labels of the two nodes are the same. Specifically, for two nodes with the same classification, the finetuned $LLM_{\text{EP}}$ is expected to output ``True", otherwise it outputs ``False".


Given that the response is limited to one token, $LLM_{\text{EP}}$ can quickly traverse the dataset's edges and reduce the number of node pairs based on the output. From the output of \( LLM_{\text{EP}} \), a new adjacency matrix \( \mathcal{A'} \) can be obtained:

$$\mathcal{A'} (i,j) = \begin{cases} 1 & \text{if } LLM_{\text{EP}}(t_i, t_j | I_{\text{EP}}) = \text{True} \\ 0 & \text{if } LLM_{\text{EP}}(t_i, t_j | I_{\text{EP}}) = \text{False} \end{cases}$$

\subsection{Dual-layer LLMs Finetuning Architecture}

Although directly utilizing LLMs to extract information for each node pair can generate customized messages for different neighbouring nodes, it does not guarantee that these messages are beneficial for downstream tasks. Considering the large number of parameters in LLMs, directly training the LLM as part of the model is impractical. For smaller-scale models (such as BERT), methods have been proposed to insert trainable graph adapter layers into the language model \cite{ribeiro-etal-2021-structural}. However, the substantial increase in complexity of modern LLMs makes directly applying such methods impractical, resulting in significant resource overhead.

Therefore, we aim to adapt LLMs to downstream tasks using efficient finetuning methods like LoRA \cite{hu2021lora}. However, LoRA requires training corpora composed of standard answers to finetune the LLMs. The dataset does not inherently provide the specific information that the LLMs should generate based on node pair data. This required corpus needs to be derived from node pair information and be relevant to downstream tasks.  The dual-layer LLMs finetuning architecture offers a more refined, goal-directed method, ensuring that the LLMs’ output can meet expectations. Experiments demonstrate the efficiency of this method, where a small amount of training data and time are sufficient for LanSAGNN to achieve SOTA performance. Specifically, the framework is organized into two layers:

\subsubsection{Knowledge Base LLM ($LLM_{\text{KB}}$)}

The Knowledge Base LLM is designed to provide corpora for the Extraction LLM. It takes the textual attributes of specific node pairs from the training set and downstream task labels as inputs, extracting information related to the downstream task answers from each node pair's textual attributes. This ensures that each output is customized based on the text attributes of the two nodes and is highly aligned with the downstream task labels. Specifically, a set \( \mathcal{V}_s \subseteq \mathcal{V}_{\text{train}} \) is obtained through sampling, for any \( v_i \in \mathcal{V}_s \) and a node \( v_j \) within its neighbourhood, the output \( t'_{ij} \) from the Knowledge Base LLM is generated through the instruction \( I_{\text{KB}} \):

\begin{equation}
    t'_{i,j} = LLM_{\text{KB}}(t_{i},t_{j},y_{i}|I_{\text{KB}})
\end{equation}

$I_{\text{KB}}$ directs the LLM to identify the reasons why node $v_i$ belongs to classification label $y_{i}$ based on the text information of node $v_i$ and its neighbor node $v_j$. This means that the LLM's output $t'_{ij}$ is extracted based on the textual information of the node pair ($v_i,v_j$), and is directly related to the downstream tasks, fully aligning with the design requirements of LanSAGNN. We denote by $\mathcal{T}'$ the collection of all $t'_{ij}$ for the sampled nodes $v_i$, which will be used in subsequent finetuning processes.

\subsubsection{Extraction LLM ($LLM_{\text{E}}$)}
Since the output $\mathcal{T}'$ from the Knowledge Base LLM is focused on the downstream tasks, it allows the Extraction LLM to use this output as finetuning corpus. This information not only captures distinct contextual relationships between two nodes but also ensures that these insights are practically beneficial for downstream tasks. By compiling $\mathcal{V}_s$, $\mathcal{T}'$ and $\mathcal{T}$ into the finetuning corpus $\mathcal{Z}$, an Extraction LLM that meets the requirements of LanSAGNN can be obtained. The input to this Extraction LLM consists of the original text of the node currently being processed and one of its connected nodes, requiring the model to determine the classification of the current node and to extract logical justifications for this classification from their connection.

Given an LLM parameterized by $\Phi$, the model is optimized for specific tasks using a parameter increment ($\Theta$) that is significantly smaller than $\Phi$, to obtain the Extraction LLM within the dual-layer finetuning architecture. The finetuning of the LLM using the LoRA method is implemented as:
$$  
\max_\Theta \sum_{(t_{i,j},I_\text{E},t'_{i,j})\in \mathcal{Z}} \sum_{s=1}^{|t'_{i,j}|} \log({P_{\Phi_{0}+\Delta \Phi (\Theta)}} ( \underset {s}{t'_{i,j}}|t_{i,j},I_{\text{E}},\underset{<s}{t'_{i,j}}  ) ),
$$
 $t_{i,j}$ represents the original text of node $v_i$ and one of its neighbouring nodes $v_j$. $I_\text{\text{E}}$ denotes the instruction used by the $LLM_\text{E}$. $P$ represents the output probability, $s$ refers to the current time step index in the sequence, typically indicating the token currently being considered or generated. $<s$ represents all sequences prior to the current time step $s$. After finetuning, for all nodes $v_{i} \in \mathcal{V}$, the output $t''_{i}$ is obtained by the $LLM_{\text{E}}$ based on the original textual information of node $v_i$ and its neighbor node $v_j$:
\begin{equation}
    t''_{i} = LLM_{\text{E}}(t_{i,j}|I_{\text{E}} )
\end{equation}

Overall, the finetuning architecture is designed to leverage graph structures and align LLMs with downstream tasks through coordinated finetuning. Experiments demonstrate that significant performance improvements can be achieved with only a small amount of data and time (See appendix).

\subsection{Downstream Pipeline}

The final step is to map the output text from the Extraction $LLM_{\text{E}}$ to the labels of downstream classification tasks. The output of the $LLM_{\text{E}}$ depends on node pairs, where the texts generated in $LLM_E$ by node $v_i$ and all its neighboring nodes $v_j \in \mathcal{N}(v_i)$ are aggregated and concatenated, forming a comprehensive text. Through an embedding model, the output of the $LLM_{\text{E}}$ can be mapped to specific downstream tasks:
\begin{equation}
    x'_i=\text{Embedding}(t_i   \underset{v_{j} \in \mathcal{N}(v_{i}) }{||}   t''_{ij})
\end{equation}

The $x'_i$ represents the embedding of node $v_i$, which serves as the node's final representation and is used as input to the GNNs for downstream tasks.

\section{Experiment}
We conduct experiments on LanSAGNN using four common TAGs datasets for node classification tasks, which include Cora \cite{mccallum2000automating}, Pubmed, Citeseer \cite{sen2008collective}, and Wiki-CS \cite{mernyei2020wiki}. These datasets encompass various domains of academic papers and Wikipedia entries, where each node represents either a paper or a Wikipedia entry, and the edges represent citation relationships between them. The experiments were carried out on two RTX A6000 GPUs, exploring various experimental setups, LLMs, and the impact of hyperparameters. Overall, this section investigates the following research questions:
\begin{itemize}
\item \textbf{RQ1}: How does LanSAGNN's performance compare to traditional GNNs and methods based on LLMs?

\item \textbf{RQ2}: What are the roles of the various components in LanSAGNN? Does using different LLMs significantly affect the model's performance?

\item \textbf{RQ3}: How do the hyperparameter settings of LanSAGNN affect the model's efficiency? What is the performance of the intervention of optional edge filters?
\end{itemize}

\begin{table*}[]
\centering
\small
\renewcommand\arraystretch{0.5}
\caption{Node classification accuracy (\%) and standard deviation under a random split of 60\% / 20\% / 20\% are reported, with experimental data averaged over ten runs.}
\begin{tabular}{c|c|c|c|c|c}
\toprule
\textbf{Category}                                                             & \textbf{Method}      & \textbf{Cora}                   & \textbf{Citeseer}                 & \textbf{Pubmed}       & \textbf{Wiki-CS}      \\

\midrule

\multirow{4}{*}{GNNs}                                                & GCN                  & 88.05 ± 0.98                    & 79.26 ± 1.45                      & 88.90 ± 0.32          & 82.81 ± 1.14            \\
                                                                     & GraphSAGE                 & 87.87 ± 1.30                     & 79.03 ± 1.03                      & 86.85 ± 0.11          & 83.76 ± 0.77            \\
                                                                     & GAT                  & 85.70 ± 0.94                    & 80.30 ± 1.35                      & 83.28 ± 0.12          & 83.42 ± 0.52            \\
                                                                     & GNN-SOTA           &  89.59 ± 1.58 & 82.07 ± 1.04 &  91.56 ± 0.50    &  84.61 ± 0.53        \\

\midrule
                                                                     
\multirow{5}{*}{\begin{tabular}[c]{@{}c@{}}LLM\\ Based\end{tabular}} & LLM Based Embedding      & 89.91 ± 0.94                    & 79.94 ± 1.20                      & 92.79 ± 0.31        & 85.63 ± 0.59            \\
                                                                     & TAPE                 & 90.22 ± 0.79                    & 81.92 ± 1.32                      & 94.33 ± 0.24          & 85.83 ± 0.30          \\
                                                                     & LLM Based A-D \& LPA & 89.16 ± 0.96                    & 76.84 ± 0.76                      & {\ul 94.75 ± 0.36}    & 86.11 ± 0.47                      \\
                                                                     & InstructGLM          & 90.77 ± 0.52                     & -                                 & 94.62 ± 0.13          & -                       \\
                                                                     & GraphEdit            & 90.90 ± 1.16                    & 81.85 ± 1.42                      & 94.09 ± 0.28          & 85.85 ± 0.64                      \\

\midrule
                                                                     
\multirow{2}{*}{Ours}                                                & LanSAGNN             & \textbf{91.24 ± 0.98}           & \textbf{82.34 ± 1.13}             & 94.69 ± 0.22          & \textbf{87.56 ± 0.44} \\
                                                                     & LanSAGNN (self loop) & {\ul 91.02 ± 0.51}              & {\ul 82.12 ± 0.95}                & \textbf{95.13 ± 0.25} & {\ul 87.34 ± 0.42}\\

                                                                     \bottomrule

\end{tabular}

\label{thetable2}
\end{table*}

\subsubsection{Main Experiment (\textbf{RQ1})}
We evaluate LanSAGNN on four TAGs datasets, using classical GNN methods such as GAT, GCN, GraphSAGE and each dataset's SOTA techniques as baseline comparisons. The GNN-SOTA in the tables refers to the best-performing methods on these datasets, including ACM-Snowball-3, ACM-Snowball-2 \cite{luan2021heterophily}, NHGCN \cite{chaudhary2024gnndld}, and CGT \cite{hoang2023mitigating}. The LLM-based baselines represent current representative methods using LLMs to address TAGs problems, each optimized graph issues from different perspectives. LLM-Embedding utilizes LLM-based embedding models in conjunction with GNNs to perform downstream tasks. TAPE instructs LLMs to generate pseudo-label rankings and corresponding rationales based on the node's original text. The texts are then embedded and input into GNNs. For fairness, the embedding models and GNNs used in the process are the same as those used in LanSAGNN. GraphEdit uses LLM finetuning to optimize the graph topology, while LLM-Based A-D \& LPA \cite{sun2023large} builds on optimizing graph structure with LLM and enhances it with TAPE’s instructions for pseudo-label propagation. InstructGLM directly inputs node texts, structural information, and neighbouring raw texts for finetuning.

Considering the high inference cost of large models, we reduce the number of edges in the datasets through random sampling to save time. Specifically, for the Cora and Citeseer datasets, we retain a maximum of two edges per node, while for the Pubmed and Wiki-CS datasets, we preserve at most one edge per node. For the dual-layer LLMs finetuning architecture, we utilize GPT-3.5-turbo as the Knowledge Base LLM and Vicuna1.5-13b \cite{zheng2024judging} as the Extraction LLM. The impact of the number of samples and different LLMs on performance is discussed later in this section (RQ3).

Experimental results, as shown in Table \ref{thetable2}, demonstrate that LanSAGNN achieve outstanding performance on these datasets. LanSAGNN (self-loop) is inspired by the traditional GNNs' self-loops and denotes the use of instructions for text enhancement of the node itself, which is then combined with text generated by the Extraction LLM. This approach aims to address the dependency of LanSAGNN on edge-generated text, indicating that text outputs cannot be obtained for isolated nodes within datasets, thereby ensuring the universality of LanSAGNN. Experiments show that on most datasets, the results of LanSAGNN (self loop) are slightly inferior to LanSAGNN. This may be due to the quality of the text enhancement based on the node itself not being as high as the output from the Extraction LLM in LanSAGNN. Due to space constraints, more detailed parameters, instructions, and other specifics will be studied in the appendix. Additionally, methods based on LLMs generally outperform traditional GNN approaches, demonstrating the potential and advantages of LLMs' contextual understanding capabilities in solving TAGs tasks. Comparisons with LLM-based methods also prove that LanSAGNN can more effectively utilize the structural information of graphs to achieve better performance.

\begin{table}[]
\centering
\small
\caption{Accuracy(\%) and standard deviation of comparison and ablation experiments}
\begin{tabular}{c|c|c}
\toprule
     \textbf{Setting}  & \textbf{Cora}       & \textbf{Citeseer}   \\  \midrule
-w/o Finetune   & 90.38 ± 0.95 & 81.24 ± 1.41 \\
-w/o ES        & 89.83 ± 0.53 & 80.05 ± 1.00 \\
-w/o OriginText & 91.03 ± 1.12 & 81.51 ± 1.22 \\ \midrule
$LLM_\text{E}$: llama3-8b     & 91.12 ± 0.85 & 81.95 ± 0.97 \\
$LLM_\text{E}$: vicuna1.5-7b  & 91.22 ± 0.87   & 82.01 ± 0.93 \\
$LLM$: llama3-8b  & 91.16 ± 0.80   & \textbf{82.54 ± 1.18} \\ \midrule
LanSAGNN      & \textbf{91.24 ± 0.98} & 82.34 ± 1.13\\ \bottomrule
\end{tabular}
\label{thetable3}
\end{table}

\subsubsection{Ablation and Comparative Experiment (\textbf{RQ2})}
We test the performance of each component of LanSAGNN through ablation studies, with results presented in Table \ref{thetable3}. The label ``w/o Finetune" indicates that the baseline LLM is not finetuned; the Extraction LLM generates text based solely on the node pair information, but this content has low relevance to the downstream tasks, resulting in a performance decrease, thereby proving the effectiveness of the dual-layer LLMs finetuning architecture. The ``w/o ES" condition utilizes graph structures only at the LLM level, and after obtaining embeddings, classification is performed using an MLP, lacking the utilization of topological structure at the embedding level compared to the complete LanSAGNN. The ``w/o OriginText" condition uses only the answers from the Extraction LLM as input for embeddings, without concatenation with the node’s original text. ``$LLM_\text{E}$: llama3-8b" and ``$LLM_\text{E}$: vicuna1.5-7b" represent replacements with different Extraction LLM. ``$LLM$: llama3-8b'' represents that both the Knowledge Base LLM and the Extraction LLM are replaced. By experimenting with open-source LLMs with various parameters and architectures, the method demonstrates its stability and versatility, showing that it does not rely on any specific  LLM.

\subsubsection{Hyperparameter and Optional Edge Filter (\textbf{RQ3})}
For LanSAGNN, the core hyperparameter is the number of edge samples per node ($k$). We experiment with different values of $k$ and the use of the Optional Edge Filter (OEF) module, as shown in Table \ref{thetable4}. The experimental results indicate that even with $k$=1, LanSAGNN maintains excellent performance. As $k$ increases, LanSAGNN's performance improves significantly. This suggests that LanSAGNN achieves better performance compared to current mainstream methods without increasing complexity while also demonstrating greater potential. The OEF module, on the other hand, reduces the model's complexity with almost no loss in performance.

Additionally, experiments also demonstrate that the dual-layer LLMs finetuning architecture requires very little time and data, and the OEF module can effectively reduce the overall time complexity (See appendix).

\begin{table}[]
\centering
\small
\caption{Accuracy(\%) and standard deviation under different $k$ values and optional edge filter (OEF).}
\renewcommand\arraystretch{0.51}
\centering
\setlength\tabcolsep{4pt} 
\begin{tabular}{c|c|c|c|c}
\toprule
\textbf{$k$}    & \textbf{Cora}         & \textbf{Citeseer}     & \textbf{Cora(OEF)}    & \textbf{Citeseer(OEF)} \\ \midrule
1  & 91.05 ± 1.01 & 81.84 ± 1.21 & - & -  \\
3  & 91.13 ± 1.03 & 82.14 ± 1.16 & 91.09 ± 0.78 & 82.21 ± 1.34  \\
5  & 91.31 ± 1.05 & 82.39 ± 1.07 & 91.24 ± 0.91 & 82.51 ± 1.33  \\
10 & 91.33 ± 0.97 & 82.67 ± 1.54 & 91.26 ± 0.76 & 82.77 ± 0.97  \\
$\infty$  & 91.37 ± 0.62 & 82.45 ± 1.27 & 91.35 ± 1.05 & 82.84 ± 1.19  \\
\bottomrule
\end{tabular}
\label{thetable4}
\end{table}

\section{Conclusion}
This work presents a novel paradigm that outperforms existing LLM-based methods by maintaining simplicity alongside a flexible and efficient finetuning architecture for the rapid optimization of LLM outputs. While LanSAGNN demonstrates outstanding performance and efficiency, it operates within the broader context of industry-wide challenges, such as interpretability issues arising from integration with LLMs and the inherently complex deployment processes associated with finetuning architectures. Addressing these prevalent obstacles will guide our future research efforts. \textbf{For more experimental results, including finetuning sampling rate tests, robustness tests, time and economic cost analysis, case studies, and more detailed experimental parameters and explanations, please refer to the appendix.} 

\bibliographystyle{IEEEbib}
\bibliography{icme2025references}

\vspace{12pt}

\end{document}